\documentclass[12 pt]{article}

\usepackage{graphicx} 
\usepackage[T1]{fontenc}

\usepackage{amsmath,amssymb}
\usepackage{mathtools}          
\usepackage{amsthm}             

\title{Credibility-Weighted Pricing of Autonomous Vehicle Liability Under Operational Design Domain Shift}
\author{Doyeon Jang}
\date{June 2026}

\renewcommand{\thesection}{\arabic{section}.}
\renewcommand{\thesubsection}{\thesection\arabic{subsection}.}
\renewcommand{\thesubsubsection}{\thesubsection\arabic{subsubsection}.}
\renewcommand{\thetable}{\arabic{table}}

\usepackage[bottom=1in, left=1in, right=1in, top=1in, includehead=false]{geometry}
\usepackage{changepage}

\usepackage{titlesec} 

\usepackage[font=small,labelfont=bf]{caption}
\usepackage{array}
\usepackage{booktabs}
\renewcommand{\arraystretch}{1.5}

\usepackage{listings} 
\usepackage{xcolor} 

\usepackage{tabularx} 

\usepackage{hyperref}
\hypersetup{
    colorlinks=true,
    urlcolor=blue,
    breaklinks=true
}

\usepackage{appendix} 

\usepackage{newtxtext,newtxmath} 

\usepackage{caption}

\makeatletter
\renewcommand{\@fnsymbol}[1]{\ensuremath{\ifcase#1\or 1\or 2\or 3\or 4\or 5\or 6\or 7\or 8\or 9\else *\fi}}
\makeatother

\titleformat{\section}
  {\normalfont\fontsize{14pt}{16pt}\selectfont\bfseries}
  {\thesection}{1em}{}
\titleformat{\subsection}
  {\normalfont\fontsize{12pt}{14pt}\selectfont\bfseries}
  {\thesubsection}{1em}{}
\titleformat{\subsubsection}
  {\normalfont\fontsize{12pt}{14pt}\selectfont\bfseries}
  {\thesubsubsection}{1em}{}

\begin{document}

\vspace*{.25em}  

\begin{center}
  {\fontsize{18pt}{22pt}\selectfont\bfseries Credibility-Weighted Pricing of Autonomous Vehicle Liability Under Operational Design Domain Shift}

  \vspace{0.8em}
  Doyeon Jang

\end{center}


{
\begin{center}
\textbf{Abstract}
\end{center}

\begin{adjustwidth}{5em}{5em}
Automated Driving System deployments create a foundational ratemaking challenge: sparse experience, shifting operational design domains, and non-stationary risk across software releases. We propose a hierarchical Bayesian credibility framework pooling across cities, software versions, and territories via a learned ODD-similarity kernel, nesting Bühlmann–Straub as a limiting case. Demonstrated on 648 verified-engaged Waymo crashes across four U.S. metros from the NHTSA Standing General Order database against 116 million matched miles, city-aggregate credibility weights are moderate (0.12–0.46), partial pooling decisively outperforms no pooling, and a power analysis shows the learned kernel's advantage becomes detectable at approximately twelve deployed cities.
\end{adjustwidth}}

\vspace{0.8em}

\noindent Keywords: Automated driving systems, credibility theory, hierarchical Bayesian models, representation learning, telematics, ratemaking, operational design domain

\section{Introduction}

\subsection{Research context}
In 2024, Waymo’s automated driving system recorded nine third-party property-damage claims and two bodily-injury claims across 25.3 million fully autonomous miles---an 88 to 92 percent reduction relative to the human-driven vehicle population in the same operating regions (Di Lillo et al., 2024b). Similar magnitudes have been reported using police-reported crash data (Kusano et al., 2024). For the actuarial profession these findings are both remarkable and unsettling. Remarkable, because they suggest the risk pool insurers price is undergoing a structural transformation. Unsettling, because the standard tools of ratemaking---credibility theory, territory relativities, loss development---were built for a world in which exposure is abundant, the insured risk is roughly stationary, and the geographic distribution of driving reflects the residence of the insured. None of these assumptions hold cleanly for ADS deployments.

\subsection{Objective}
The retrospective benchmarking literature has done valuable work establishing that ADS performance differs significantly from HDV performance and has begun to grapple with methodological pitfalls such as reporting-threshold misalignment and ODD confounding (Scanlon et al., 2024b; Chen et al., 2024). What this literature has not yet addressed is the prospective pricing problem. When an ADS operator enters Miami, or when a new operator begins commercial service in a city with no prior ADS exposure, an insurer must produce a rate. The retrospective benchmarks tell us what happened in San Francisco, Phoenix, Los Angeles, and Austin; they do not tell us how to transfer that experience to a city with different road geometry, pedestrian density, and signal phasing. The classical credibility apparatus reacts to the volume of own experience: when a city has accumulated enough events its own rate is credible, and when it has not, the estimate shrinks toward the portfolio mean. The novel difficulty in the ADS setting is that even when aggregate own experience is substantial, it is concentrated in a handful of cities, and the cells that actually drive a rate---a given software version, in a given period, in a given territory---remain sparse.

This paper proposes a framework for that transfer problem. We combine hierarchical Bayesian credibility with a learned ODD-similarity metric drawn from the representation-learning literature in machine learning. The similarity metric is trained on publicly available geospatial and HDV claims data and is used to govern how credibility flows between cities, software versions, and territories within a city. We show that the resulting framework nests classical Bühlmann–Straub as a special case while gaining the flexibility to incorporate the richer covariate structure that ADS pricing demands, and we demonstrate empirically that it produces sensible, interpretable, and well-calibrated posterior predictions for new deployments.

The contributions of this paper are threefold. First, we formalize the ADS pricing problem as a hierarchical partial-pooling problem over cities, ODDs, and software versions, and derive a posterior credibility weight that reduces to the classical Bühlmann–Straub form in the limit; the full derivation appears in Appendix B. Second, we introduce a learned ODD-similarity metric, constructed via supervised contrastive learning over geospatial road-network and exposure features, and integrate it into the hierarchical prior as a Gaussian-process kernel governing cross-territory credibility flow; the posterior predictive distribution for an unobserved city becomes a conditional Gaussian whose moments depend explicitly on the city’s similarity to deployed cities (Appendix C). Third, we demonstrate the framework empirically on NHTSA SGO data, quantify the SGO-to-liability reporting-threshold gap, and produce prospective frequency estimates for hypothetical new deployments in Miami, Boston, and Denver.

\subsection{Outline}
The remainder of the paper is organized as follows. Section 2 reviews classical credibility theory, the ADS safety benchmarking literature, and recent developments at the intersection of machine learning and actuarial science. Section 3 formalizes the prospective pricing problem. Section 4 develops the hierarchical credibility model and derives its reduction to Bühlmann–Straub. Section 5 introduces the learned ODD-similarity metric and its integration with the credibility model. Section 6 presents the empirical demonstration, including a comparison of the proposed framework against baselines via leave-one-city-out predictive log-likelihood. Section 7 discusses implications for ratemaking, reserving, and the regulatory landscape. Section 8 concludes. Appendices contain the NumPyro implementation, the Bühlmann–Straub limit derivation, the conditional Gaussian for the new-city posterior, and the data-assembly protocol used in the empirical section.

\section{Background and methods}

\subsection{Classical Credibility Theory}
Credibility theory provides the actuarial foundation for combining a unit's own claim experience with a portfolio mean when own experience is insufficient. The Bühlmann (1967) model assumes a portfolio of risk units indexed by $i = 1, ..., K$, each observed over $T_i$ periods with exposure $w_{it}$ and pure premium $X_{it}$. Each unit has a latent risk parameter $\theta_i$ drawn from a structure distribution, and the credibility estimator for unit $i$'s expected pure premium takes the form of a weighted average between own experience and the grand mean. The Bühlmann-Straub (1970) extension accommodates heterogeneous exposure across units, yielding a credibility weight 

\noindent\makebox[\textwidth][s]{\hfill $Z_i = w_{i\cdot} / (w_{i\cdot} + K)$, where $K = \operatorname{E}[\sigma^2(\theta)] / \operatorname{Var}[\mu(\theta)]$ \hfill (2.1)}

with $w_{i\cdot} = \sum_t w_{it}$. More exposure pulls $Z_i$ toward 1, greater between-unit variance relative to within-unit variance pulls $Z_i$ toward 1, and everything else shrinks toward the grand mean.

Modern treatments by Klugman, Panjer, and Willmot (2012) and Bühlmann and Gisler (2005) generalize this structure to hierarchical models, where risk units are grouped into larger collectives and credibility weights are derived at each level of the hierarchy. The problem for ADS pricing, however, is that the standard hierarchical credibility model treats units within a level as exchangeable. A Phoenix surface-street mile and a hypothetical Miami surface-street mile receive the same shrinkage, even though their underlying ODDs differ substantially in pedestrian density, intersection geometry, and weather exposure. We want a credibility weight that depends on which other units we are borrowing from.

A second limitation of the classical apparatus, in the ADS context, is that it presumes the latent risk parameter $\theta$ is stationary. In ratemaking for human-driven vehicles this presumption is approximately valid: the driving population evolves only gradually. For ADS, by contrast, each software release is a discontinuous change in the insured risk. The Waymo Driver of mid-2023 is not the same risk as the Waymo Driver of mid-2024, and the classical credibility apparatus has no native mechanism to express that fact. We therefore augment the hierarchical structure with explicit software-version random effects in Section 4.

\subsection{ADS Safety Benchmarking}
The retrospective ADS safety benchmarking literature has matured rapidly. Early work by Favarò et al. (2017) analyzed California disengagement and collision reports for autonomous test vehicles. Chen and Shladover (2024) compared crash rates within San Francisco across four populations: Uber rideshare, supervised AVs, driverless Waymo, and driverless Cruise. Kusano et al. (2024) extended this comparison to police-reported crash databases over 7.1 million Waymo miles, applying corrections for vehicle type, road type, and underreporting, and reported a 55 percent reduction in police-reported crash rates and an 80 percent reduction in any-injury-reported crash rates.

Di Lillo et al. (2024a) introduced the use of third-party liability insurance claims as a comparative safety measure. Over an initial 3.8 million autonomous miles, they observed substantial reductions in both property-damage and bodily-injury claim frequencies relative to a zip-code-calibrated HDV benchmark. Di Lillo et al. (2024b) extends this analysis to 25.3 million miles across San Francisco, Phoenix, Los Angeles, and Austin, and introduces a latest-generation HDV benchmark capturing vehicles equipped with modern advanced driver assistance systems (ADAS). The Waymo ADS outperformed both benchmarks with statistical significance: 88 and 92 percent reductions in property-damage and bodily-injury claims relative to the overall driving population, and 86 and 90 percent relative to the latest-generation HDV benchmark.

Waymo’s publicly reported mileage milestones provide the exposure base for the present demonstration. The operator’s Safety Impact disclosures report 170.7 million rider-only miles through December 2025 across the four deployed metros and the 200-million-mile milestone in February 2026 (Waymo, 2026). We use these cumulative disclosures, interpolated to the SGO observation window, to construct exposure denominators for the SGO crash counts (Section 6.1).

\subsection{Machine Learning in Actuarial Science}

The integration of machine learning into ratemaking and reserving has accelerated over the past decade. Wüthrich and Buser (2025) give a rigorous treatment of statistical foundations of actuarial learning, including generalized linear models, neural networks, and the hybrid GLM–neural-network proposed by Wüthrich and Merz (2019). Embedding-based representations of categorical features have proven effective\textemdash Richman (2021) for mortality, and Delong et al. (2021) for developing gamma mixture density networks for claim-amount modelling.

Within this literature, the closest precedent for our approach is the use of learned embeddings to define actuarial territories. Traditional territory definitions rely on hand-crafted aggregations (zip codes, counties, rating territories) that may not capture the latent structure of risk. Learned embeddings, by contrast, place each geographic unit in a vector space where distance reflects risk similarity. We extend this to ADS pricing by training the embedding on ODD-relevant features---road geometry, intersection density, pedestrian exposure---and using the similarity metric to govern credibility flow in a Bayesian hierarchical model. 

Our contrastive objective draws on SimCLR (Chen et al., 2020) but, in the spirit of Khosla et al. (2020), defines positive pairs by similarity of an auxiliary outcome (HDV claim frequency) rather than by data augmentation; the resulting embedding serves as the input to a kernel that parameterizes a Gaussian-process prior on random effects.

\section{Problem Formulation}

Consider an insurer pricing third-party liability coverage for an ADS deployment. Let $c$ index city (or, more generally, operating region), $v$ index ADS software version, and $t$ index time period. Let $N_{c,v,t}$ denote the count of third-party liability claims and $E_{c,v,t}$ the exposure measured in millions of autonomous miles. Following standard practice in collision frequency modeling, we assume 

\noindent\makebox[\textwidth][s]{\hfill $N_{c,v,t} \mid \lambda_{c,v,t} \sim \mathrm{Poisson}(\lambda_{c,v,t} \cdot E_{c,v,t})$ \hfill (3.1)}

where $\lambda_{c,v,t}$ is the expected claim frequency per million miles. The pricing problem is to produce a point estimate or posterior distribution for $\lambda_{c^*,v^*,t^*}$ at a target deployment $(c^*, v^*, t^*)$, where direct experience may be zero or near-zero.

Three features distinguish this problem from standard ratemaking. First, exposure is concentrated: in the present demonstration the SGO window carries roughly 116 million four-city rider-only miles, but more than 90 percent of those miles fall in just three metros, and within a metro they are spread thinly across software versions and quarters. Second, the operational design domain shifts as fleets expand: a new-city deployment is not a draw from the same risk distribution as existing cities. Third, the insured risk itself is non-stationary: each software update potentially modifies the underlying claim frequency, breaking the assumption of a stable $\theta$ in classical credibility theory.

The actuary's task is to construct $\hat\lambda_{c^*,v^*,t^*}$ by borrowing strength from three sources: (a) the operator's own ADS experience in other cities and software versions, (b) HDV claim experience in territories similar to $c^*$, and (c) the latest-generation HDV experience capturing ADAS-equipped vehicles, which proxies for the technological frontier of the human-driven fleet. The framework developed in Sections 4 and 5 makes this borrowing explicit and principled.

\section{Hierarchical Credibility Model}

\subsection{Model Specification}

We specify a hierarchical Bayesian Poisson generalized linear model with random effects at the city, software-version, and city-version interaction levels:

\noindent\makebox[\textwidth][s]{\hfill $N_{c,v,t} \mid \lambda_{c,v,t} \sim \mathrm{Poisson}(\lambda_{c,v,t} \cdot E_{c,v,t})$ \hfill (4.1)}

\noindent\makebox[\textwidth][s]{\hfill $\log \lambda_{c,v,t} = \beta_0 + x_{c,v,t}^T \beta + \alpha_c + \gamma_v + \delta_{c,v}$ \hfill (4.2)}

The fixed-effect covariates $x_{c,v,t}$ capture observable ODD attributes: road class mix (proportion of arterial, collector, and local-road miles), intersection density per square kilometer, signalized-intersection fraction, weather exposure (proportion of operating hours in rain or fog), and time-of-day mix of operating hours. The random effects encode the hierarchy:

\noindent\makebox[\textwidth][s]{\hfill $\alpha_c \sim N(0, \tau_c^2), \quad \gamma_v \sim N(0, \tau_v^2), \quad \delta_{c,v} \sim N(0, \tau_{cv}^2)$ \hfill (4.3)}

The hyperpriors on the standard deviations require care. A standard weakly-informative default in the hierarchical-model literature is Half-Normal(0,1) on each scale. In the ADS regime this default proved too diffuse, producing posterior predictive intervals spanning several orders of magnitude for hypothetical new cities. We therefore use the moderately tighter

\noindent\makebox[\textwidth][s]{\hfill $\tau_c, \tau_v \sim \mathrm{Half-Normal}(0,0.5), \quad \tau_{cv} \sim \mathrm{Half-Normal}(0, 0.3)$ \hfill (4.4)}

The tighter prior on the interaction $\tau_{cv}$ reflects the prior expectation that city-by-version effects are smaller than city or version main effects\textemdash a standard prior-elicitation judgement in hierarchical models with crossed random effects, where interaction variance components are routinely initialized below main-effect scales absent domain knowledge to the contrary. The city random effect $\alpha_c$ captures residual city-level risk after controlling for observed ODD covariates. The software-version effect $\gamma_v$ captures non-stationarity in the ADS itself: an updated planner, perception stack, or behavioral policy may shift the claim-frequency baseline. The interaction $\delta_{c,v}$ permits certain software versions to perform differently in certain ODDs---a meaningful possibility, since a stack trained predominantly on Phoenix surface-street data may underperform when first deployed in the denser intersection grid of San Francisco. The fixed-effect coefficient vector $\beta$ receives a weakly informative prior $\beta \sim N(0, 0.5^2 \cdot I)$, following Gelman et al.'s (2008) standard recommendation for default priors on logistic and Poisson GLM coefficients. The intercept $\beta_0$ receives a wider $\mathrm{N}(0, 2.5^2)$ prior on the log-frequency scale. 

\subsection{Posterior Inference}

Posterior inference proceeds via the No-U-Turn Sampler (NUTS), an adaptive variant of Hamiltonian Monte Carlo, in NumPyro (Phan et al., 2019); the full implementation is in Appendix A. The non-centered parameterization of the random effects---standard practice for hierarchical models with small group sizes---accelerates mixing and avoids the funnel pathologies described by Betancourt and Girolami (2015). For random effect $\alpha_c$, the non-centered parameterization replaces the joint sampling of $(\tau_c, \alpha_c)$ with the equivalent reparameterization

\noindent\makebox[\textwidth][s]{\hfill $\tilde{\alpha}_c \sim N(0, 1), \quad \alpha_c = \tau_c \cdot \tilde{\alpha}_c$ \hfill (4.5)}

and analogously for $\gamma_v$ and $\delta_{c,v}$. This trick removes the strong posterior correlation between the random effect and its scale parameter that would otherwise frustrate NUTS's leapfrog integrator.

\subsection{Reduction to Bühlmann-Straub}

The hierarchical model nests Bühlmann-Straub as a limiting case. The full derivation appears in Appendix B; we sketch the key steps here. Consider the simplified model with no covariates, no software-version effect, and no interaction: 

\noindent\makebox[\textwidth][s]{\hfill $\log \lambda_c = \beta_0 + \alpha_c, \quad \alpha_c \sim N(0, \tau^2)$ \hfill (4.6)}

Under a Gaussian approximation to the Poisson likelihood (valid when expected counts are moderate or when one applies a Laplace approximation around the posterior mode), the posterior mean of $\alpha_c$ admits the closed-form expression

\noindent\makebox[\textwidth][s]{\hfill $\mathrm{E}[\alpha_c \mid \text{data}] = Z_c \cdot (\log \hat{\lambda}_c^r - \beta_0)$ \hfill (4.7)}

where $\hat{\lambda}_c^r = N_c / E_c$ is the maximum-likelihood estimate from city $c$'s own experience, and the credibility weight is 

\noindent\makebox[\textwidth][s]{\hfill $Z_c = (E_c \cdot \hat{\lambda}_c^r \cdot \tau^2) / (E_c \cdot \hat{\lambda}_c^r \cdot \tau^2 + 1)$ \hfill (4.8)}

This has the same shape as the Bühlmann-Straub weight $Z = w/(w+K)$, with effective exposure $E_c \cdot \hat{\lambda}_c^r$ playing the role of $w$ and $1/\tau^2$ playing the role of K. 

\section{ODD Similarity via Learned Embeddings}

\subsection{Motivation}

The hierarchical model of Section 4 treats cities as exchangeable conditional on observed covariates. In practice, two cities may share similar observed covariates yet differ on latent features that affect claim frequency---pedestrian behavior patterns, signal-timing conventions, and road-user mix. Conversely, two cities may differ on individual observed covariates yet be functionally similar from the standpoint of ADS risk. The first case generates a false signal; the second wastes signal. We address both by introducing a learned ODD-similarity metric and using it to define a Gaussian-process prior on the city random effects.

Why learn the similarity rather than specify it? Because the appropriate notion of similarity for ADS risk is not obvious a priori. Geographic distance is a poor proxy: downtown Phoenix and suburban Phoenix are geographically close but ODD-dissimilar, while downtown Phoenix and downtown Austin are far apart yet share substantial functional similarity in grid structure, signal density, and pedestrian patterns. Hand-crafted feature distances (Euclidean distance in the raw feature space) are sensitive to feature scaling and ignore the latent structure that the data could reveal. A learned representation can discover the right combination of features through the structure of an auxiliary task. We adopt this approach.

\subsection{Embedding Architecture}

We work at the level of H3 hexagonal cells (Uber, 2018) at resolution 8, which yield cells approximately 460 meters across. For each cell $j$, we construct a feature vector $f_j$ from publicly available sources:

\begin{itemize}
    \item Road network features from OpenStreetMap: total road length by functional class, intersection count, signalized-intersection fraction, average road segment length, and a measure of network betweenness centrality.
    \item Pedestrian and cyclist exposure proxies: population density and commute-mode share from the American Community Survey, and bicycle/pedestrian crash density from FARS aggregated to the cell.
    \item Land-use features: residential, commercial, and mixed-use share derived from OpenStreetMap land-use polygons.
    \item Historical HDV crash density: per-mile crash frequency by severity, aggregated from NHTSA FARS crash records over 2018-2022.
\end{itemize}

We train an embedding function $\phi : f_j \mapsto \mathbb{R}^d$ (with $d = 32$) using a contrastive objective. Our objective adapts the contrastive loss function of Chen et al. (2020) to a label-supervised setting in the spirit of Khosla et al. (2020): rather than constructing positive pairs by data augmentation (the SimCLR convention), we construct them by similarity of an auxiliary observed outcome (HDV claim frequency, in our case). Formally, for a batch of cells $\{j_1, ..., j_B\}$ with HDV frequencies $\{y_1, ..., y_B\}$, the loss is

\noindent\makebox[\textwidth][s]{\hfill $\mathcal{L} = -\sum_i \log \dfrac{\exp(\mathrm{sim}(\phi(f_i), \phi(f_{p(i)}))/T)}{\sum_{k \neq i} \exp(\mathrm{sim}(\phi(f_i), \phi(f_k))/T)}$ \hfill (5.1)}

where $\mathrm{sim}(\cdot, \cdot)$ denotes cosine similarity, $T$ is a temperature parameter ($T = 0.1$), and $p(i)$ selects a positive partner for cell $i$\textemdash drawn uniformly from cells whose HDV claim frequency lies within 10 percent of $y_i$. The architecture for $\phi$ is a three-layer multilayer perceptron with ReLU activations and a final L2-normalization layer. The encoder has 64 hidden units in each of its first two layers and produces a 32-dimensional embedding; the total parameter count is small enough ($\sim$6,800 parameters) to train comfortably on a few thousand H3 cells without overfitting. We train with the Adam optimizer at a learning rate of $3 \times 10^{-3}$ with weight decay $10^{-5}$, for 30 epochs and batch size of 256.

On the data run (OSM/ACS/FARS features for 5,239 H3 cells across seven cities) the supervised contrastive embedding achieves meaningful city separation. Figure 1 projects the learned cell embeddings onto the two principal axes of between-city variance, visualizing the ODD structure the contrastive objective has captured. The participation ratio of the embedding covariance eigenvalues is 4.75, and the first two principal components together account for 53.5 percent of variance (PC1 alone for 39.0 percent), reflecting a well-distributed 32-dimensional representation. The learned kernel shows moderate correlation with the one-dimensional HDV-frequency kernel (pairwise Pearson $r = 0.58$ across the 21 city pairs) and with the Euclidean kernel on raw covariates ($r = 0.64$); both are well below 1.0, confirming the contrastive objective captures ODD structure partially distinct from either baseline while sharing some coarse ordering. By contrast, the HDV-only and Euclidean kernels are nearly collinear with one another ($r = 0.95$).

\begin{figure}
    \centering
    \includegraphics[width=0.7\linewidth]{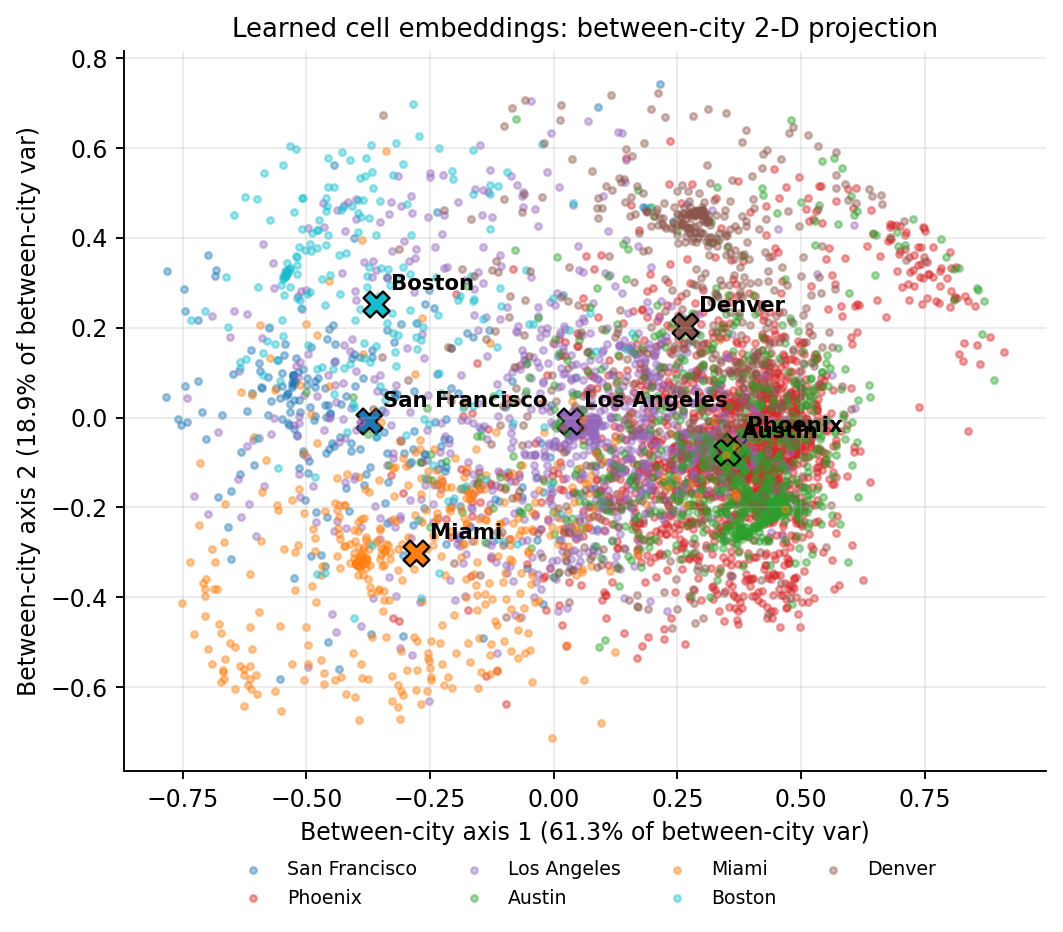}
    \caption{Projection of the learned cell embeddings.}
    \label{fig:fig1}
\end{figure}

After training, each cell, and after aggregation each city, inherits an embedding vector. We define the city-level embedding as the cell-area-weighted mean embedding for cells within the city’s extent, re-normalized to unit length so that cosine distance is well-defined.

\subsection{Integration with the Credibility Model}

Given the trained embedding, we define the ODD-similarity matrix $S$ with entries 

\noindent\makebox[\textwidth][s]{\hfill $S_{c,c'} = \exp(-\|\phi(c) - \phi(c')\|^2 / 2\ell^2)$ \hfill (5.2)}

where $\phi(c)$ is the cell-area-weighted mean embedding for city $c$ and $\ell$ is a length-scale hyperparameter inferred from the data. We choose $\ell$ so that the median pairwise similarity across cities equals 0.5---a heuristic that places the kernel’s effective bandwidth at the scale of the data---yielding $\ell^2 \approx 0.685$ in our empirical setting. We replace the independent prior on $\alpha_c$ in Section 4 with a Gaussian-process prior:

\noindent\makebox[\textwidth][s]{\hfill $\alpha = (\alpha_1, \cdots, \alpha_C)^T \sim \mathcal{N}(\mathbf{0}, \sigma^2 \cdot S)$ \hfill (5.3)}

Under this prior, cities with similar ODD embeddings have correlated random effects. Posterior inference proceeds as before; the only change is that the covariance structure of the random effects is no longer diagonal. We implement the prior via a non-centered Cholesky parameterization: 

\noindent\makebox[\textwidth][s]{\hfill $z \sim N(0,I), \quad \alpha = \sigma \cdot L \cdot Z, \quad LL^T = S$ \hfill (5.4)}

which preserves the funnel-avoidance benefits of the non-centered parameterization from Section 4.2. The scale parameter $\sigma$ plays the same role as $\tau_c$ did in Section 4\textemdash it is the marginal standard deviation of the city random effect, now under a multivariate-normal rather than independent-Normal prior\textemdash and receives the same Half-Normal(0, 0.5) prior. The remaining random effects $\gamma_v$ and $\delta_{c,v}$ retain their independent-Normal priors.

For a new deployment $c^*$, the posterior of $\alpha_{c^*}$ is obtained as a conditional Gaussian draw given the posterior of the deployed-city $\alpha$ vector. Specifically, if we let $S_{\text{dep}}$ denote the (C $\times$ C) similarity matrix over deployed cities, $s$ is the (C $\times$ 1) vector of similarities of $c^*$ to deployed cities, and $s_* = S_{c^*,c^*}$, then for each MCMC draw of $(\alpha_\text{dep}, \sigma)$ we draw

\noindent\makebox[\textwidth][s]{\hfill $\alpha_{c^*} \mid \alpha_{\text{dep}}, \quad \sigma \sim N\left( s^T S_{\text{dep}}^{-1} \alpha_{\text{dep}}, \, \sigma^2 (s_* - s^T S_{\text{dep}}^{-1} s) \right)$ \hfill (5.5)}

This is the standard conditional-Gaussian formula for a partition of a multivariate normal vector. The full derivation appears in Appendix C. The conditional mean is a weighted combination of the deployed-city $\alpha$ values, with weights set by the kernel regression coefficients $s^TS_{\text{dep}}^{-1}$; the conditional variance shrinks to zero as $c^*$ becomes maximally similar to the deployed set, and grows to $\sigma^2$ as $c^*$ becomes orthogonal to it. 

The effective credibility weight for the new deployment now takes the form 

\noindent\makebox[\textwidth][s]{\hfill $Z_{c^*} = f(E_{c^*}, \{S_{c^*, c}: c \in \text{deployed cities}\}, \sigma^2)$ \hfill (5.6)}

a function not just of own exposure but of how similar the new ODD is to ones with existing experience. When the new city resembles Phoenix in embedding space, the posterior borrows heavily from Phoenix experience. When it resembles nothing in the training set, $Z_{c^*}$ stays near zero and the model defers to the cross-city covariate prior. This is the central mathematical contribution of the framework. 

\section{Empirical Demonstration}

\subsection{Data Sources}

The empirical demonstration uses two public-data layers---an ADS event numerator and an exposure denominator---together with the ODD-similarity feature layer of Section 5.

\textbf{ADS crash counts (NHTSA SGO).} The ADS numerator is drawn from NHTSA’s Standing General Order 2021-01 incident-report file (NHTSA, 2026). We retain Waymo LLC reports whose ADS engagement status is “Verified Engaged,” deduplicate to the latest report version per incident identifier, and map the incident municipality to one of four study metros: San Francisco, Phoenix, Los Angeles, and Austin. The deployed set is restricted to these four metros, which have stable coverage across the full window and reliable exposure denominators; verified-engaged Waymo crashes in other markets that appear in the SGO file (Atlanta, plus smaller counts in Dallas, Houston, Washington, Orlando, San Antonio, Nashville, and Philadelphia) are excluded, as is Miami, whose crashes appear only in a partial window with no stable exposure denominator and which we instead retain as a prospective city in Section 6.4. After canonicalizing typographical variants of the version strings we observe three distinct software versions---5th Generation ADS Version 9, 5th Generation ADS Version 10, and 6th Generation ADS Version 10. Per-metro crash counts are San Francisco 254, Phoenix 134, Los Angeles 186, and Austin 74, for a total of 648 crashes spanning June 2025 through April 2026 across five quarterly periods. 

\textbf{ADS exposure (operator disclosure).} The SGO file records crash events but not vehicle-miles. We derive exposure denominators from Waymo’s published milestones: 170.7 million rider-only miles through December 2025 across the four deployed metros, and the 200-million-mile milestone in February 2026 (Waymo, 2026). Interpolating cumulative miles at the window boundaries and allocating by each metro’s share of Waymo’s published cumulative rider-only miles through December 2025 (San Francisco 31.4 percent, Phoenix 40.2 percent, Los Angeles 22.2 percent, Austin 6.3 percent), we estimate approximately 116 million four-city rider-only miles accumulated during the SGO window, allocated as San Francisco 36.37, Phoenix 46.62, Los Angeles 25.72, and Austin 7.29 million miles. 

\textbf{ODD-similarity features.} The cell-feature layer---OpenStreetMap road-network geometry via the Overpass API, Census ACS 5-year demographics spatially interpolated onto H3 cells, and NHTSA FARS crash records (2018–2022)---is built from public sources via the fetcher pipeline of Appendix D. It yields 5,239 H3 resolution-8 cells across seven cities. Figure 2 shows the ADS exposure distribution. Table 1 summarizes the SGO crash counts, window-matched ADS exposure, and implied crash frequencies for the four deployed metros. 

\begin{figure}[htbp]
    \centering
    \includegraphics[width=0.8\linewidth]{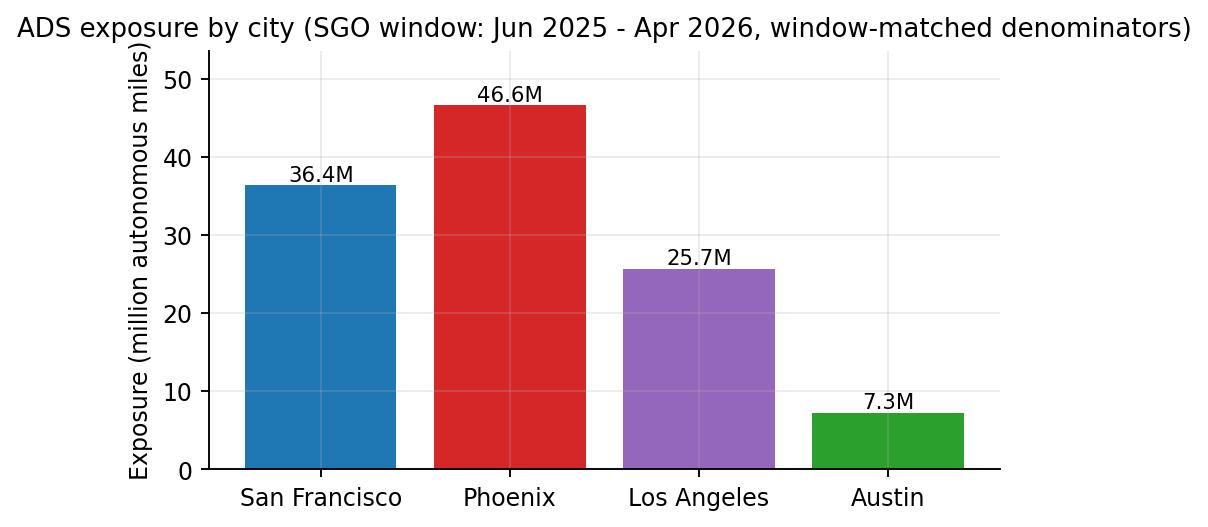}
    \caption{ADS rider-only exposure by city in the SGO observation window.}
    \label{fig:fig2}
\end{figure}

\begin{table}[h!]
\centering
\caption{SGO crash counts and implied frequencies}
\renewcommand{\arraystretch}{1.2} 
\begin{tabular}{lccc}
\toprule
\textbf{City} & \textbf{SGO crashes} & \textbf{ADS miles (M)} & \textbf{ADS freq} \\ 
\midrule
San Francisco & 254 & 36.37 & 6.98 \\ 
Phoenix       & 134 & 46.62 & 2.90 \\ 
Los Angeles   & 186 & 25.72 & 7.23 \\ 
Austin        & 74  & 7.29  & 10.15 \\ 
\midrule
\textbf{Overall} & \textbf{648} & \textbf{116.0} & \textbf{5.59} \\ 
\bottomrule
\end{tabular}
\end{table}

\subsection{Posterior Estimates Under the Hierarchical Model}

We fit two models to the SGO data: the independent-random-effects model of Section 4 (indep-RE), in which $\alpha_c \sim \text{N}(0, \tau_c^2)$, and the GP-prior model of Section 5.3 (GP-learned), in which $\alpha \sim \text{N}(0, \sigma^2\text{S})$ with S the learned ODD-similarity matrix. Both models use 2 chains of NUTS with 1,000 warm-up and 1,500 sampling iterations per chain, target acceptance probability 0.95, and the priors specified in Sections 4.1 and 5.3. 

\textbf{Convergence diagnostics.} Both models converge cleanly: the worst $\hat{R}$ is 1.00 to two decimal places for both, and the minimum bulk effective size for the GP-prior model is 1,735 of 3,000 draws. The non-centered Cholesky parameterization of the city random effects is essential: a naive centered implementation produces the funnel pathology of Betancourt and Girolami (2015). Table 2 reports diagnostics for selected parameters.

\begin{table}[htbp]
\centering
\caption{Bayesian Model Parameter Estimates}
\label{tab:model_estimates}
\begin{tabular}{lcccc}
\toprule
\textbf{Parameter} & \textbf{Post. mean} & \textbf{Post. SD} & \textbf{ESS (bulk)} & $\boldsymbol{\hat{R}}$ \\
\midrule
$\beta_0$ & $0.02$ & $0.75$ & $1,735$ & $1.00$ \\
$\beta$[arterial share] & $-0.12$ & $0.47$ & $3,230$ & $1.00$ \\
$\beta$[intersection dens.] & $-0.07$ & $0.46$ & $3,506$ & $1.00$ \\
$\beta$[signalized frac.] & $-0.38$ & $0.34$ & $2,987$ & $1.00$ \\
$\beta$[weather rain] & $0.20$ & $0.41$ & $3,240$ & $1.00$ \\
$\beta$[weather fog] & $-0.09$ & $0.43$ & $3,018$ & $1.00$ \\
$\beta$[night share] & $-0.59$ & $0.37$ & $2,863$ & $1.00$ \\
$\sigma_c$ (GP city scale) & $0.38$ & $0.29$ & $2,516$ & $1.00$ \\
$\tau_v$ (version scale) & $1.02$ & $0.26$ & $2,640$ & $1.00$ \\
$\tau_{cv}$ (interaction scale) & $0.28$ & $0.19$ & $1,798$ & $1.00$ \\
$\gamma$[5G v10, dominant] & $1.81$ & $0.70$ & $1,910$ & $1.00$ \\
$\gamma$[5G v9] & $-0.82$ & $0.70$ & $2,396$ & $1.00$ \\
$\gamma$[6G v10] & $-0.96$ & $0.79$ & $2,779$ & $1.00$ \\
$\alpha$[San Francisco] & $-0.02$ & $0.44$ & $2,943$ & $1.00$ \\
$\alpha$[Phoenix] & $-0.01$ & $0.45$ & $2,935$ & $1.00$ \\
$\alpha$[Los Angeles] & $0.09$ & $0.43$ & $2,991$ & $1.00$ \\
$\alpha$[Austin] & $0.07$ & $0.46$ & $2,716$ & $1.00$ \\
\bottomrule
\end{tabular}
\end{table}

\textbf{Bühlmann-Straub sanity check.} Before discussing the hierarchical posterior we compute the classical Bühlmann–Straub credibility weights from the city totals (Table 3), estimating the within-city process variance empirically from the (version, period) cells that make up each city rather than from a Poisson plug-in. The weights Z range from 0.12 (Austin, the sparsest city at 7 million miles) to 0.46 (Phoenix, the largest at 47 million miles). Each city’s estimate is a genuine blend of its own SGO rate and the grand mean of 5.59. The remaining shrinkage is exactly what the hierarchical model of Sections 4–5 then refines through covariates and the ODD-similarity kernel.

\begin{table}[htbp]
\centering
\caption{Classical Bühlmann-Straub credibility weights from SGO city totals}
\label{tab:city_crash_rates}
\begin{tabular}{lccccc}
\toprule
\textbf{City} & \textbf{Crashes} & \textbf{Miles (M)} & \textbf{Own rate} & \textbf{Z (BS)} & $\boldsymbol{\hat{\lambda}}$ \\
\midrule
San Francisco & $254$ & $36.37$ & $6.98$  & $0.394$ & $6.14$ \\
Phoenix       & $134$ & $46.62$ & $2.90$  & $0.455$ & $4.37$ \\
Los Angeles   & $186$ & $25.72$ & $7.23$  & $0.315$ & $6.11$ \\
Austin        & $74$  & $7.29$  & $10.15$ & $0.115$ & $6.12$ \\
\bottomrule
\end{tabular}
\end{table}

The individual (city, version, period) cells are thin---the dominant cell, 5th-Generation Version 10 in Phoenix in a single quarter, contains only a few dozen crashes---and it is precisely across these cells that the credibility framework pools using the ODD-similarity structure. Figure 3 contrasts the closed-form city-aggregate weights with the per-cell hierarchical weights. High-volume cells earn near-full own-experience credibility while sparse version cells (e.g. 5G Version 9, 6G Version 10) are shrunk heavily toward the pool.

\begin{figure}
    \centering
    \includegraphics[width=0.8\linewidth]{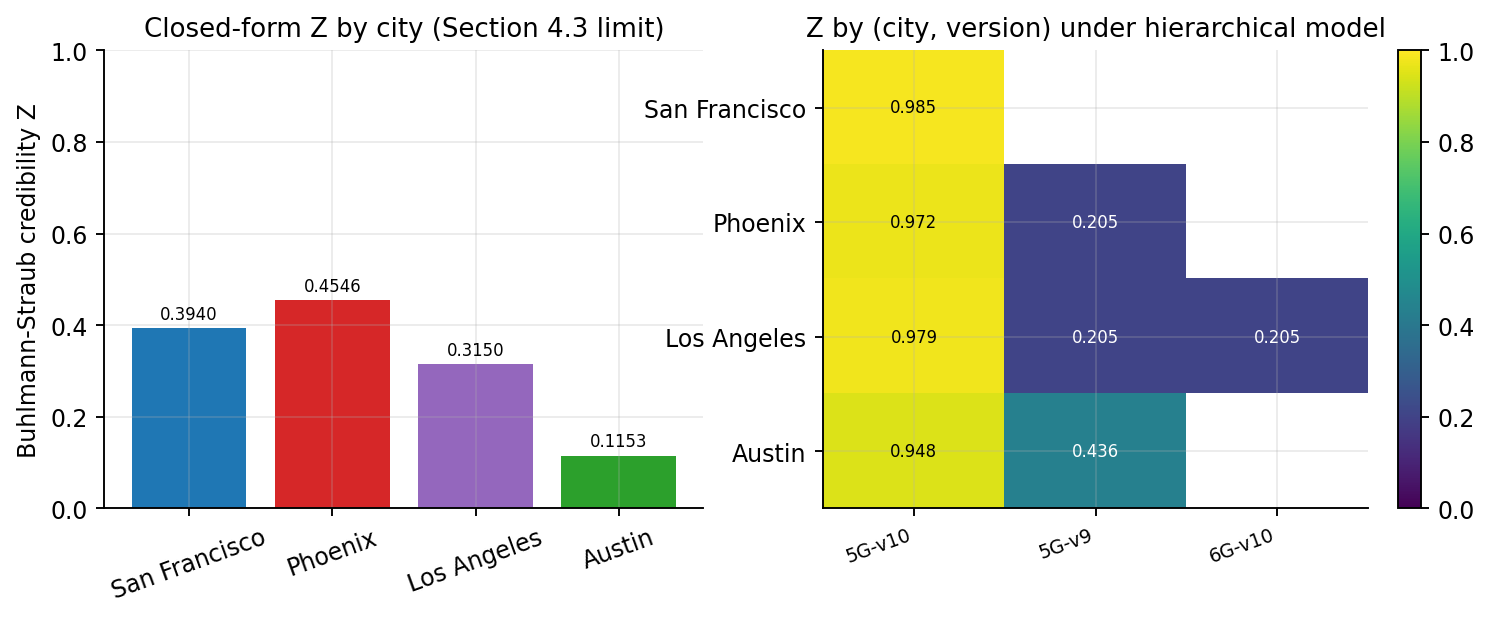}
    \caption{Left: classical Bühlmann–Straub credibility weights by city. Right: hierarchical credibility weights at posterior-mean hyperparameters by (city, version) cell.}
    \label{fig:fig3}
\end{figure}

\textbf{Posterior distributions and the learned similarity matrix.} Figure 4 visualizes the learned ODD-similarity matrix, computed from the contrastive embedding of OSM, ACS, and FARS features and aggregated to the city level. Two clusters are clearly visible. Cluster A (dense, walkable, high crash density): Boston and San Francisco at S = 0.87, sharing high intersection density, high pedestrian commute share, and high FARS crash density; Miami is adjacent to this cluster at S = 0.80 to San Francisco. Cluster B (sunbelt arterial grid, lower pedestrian share): Austin (0.91 to Phoenix), Denver (0.74 to Phoenix, 0.76 to Austin), and Phoenix at mutual similarities S $\in$ [0.74, 0.91]. Los Angeles bridges the clusters, with similarities of 0.54 to San Francisco, 0.60 to Austin, and 0.53 to Phoenix. Miami's position near the dense-urban cluster reflects its higher intersection density and higher pedestrian commute share.

\begin{figure}
    \centering
    \includegraphics[width=0.7\linewidth]{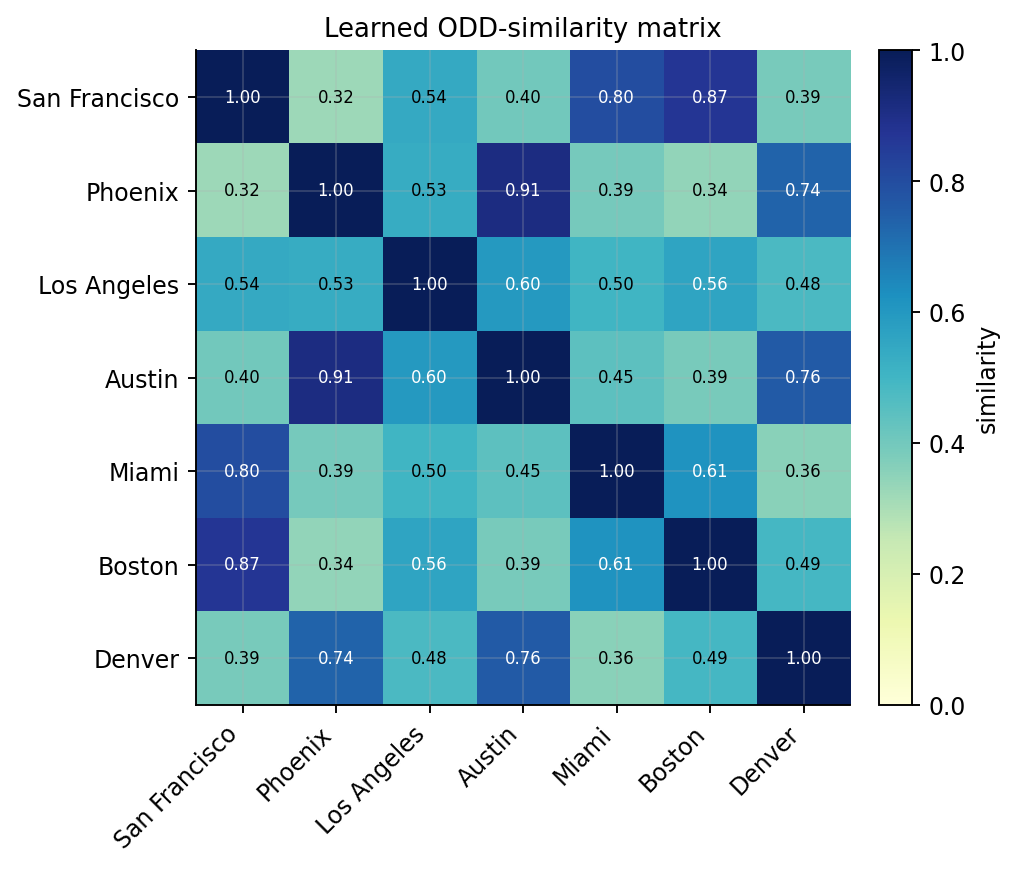}
    \caption{Learned ODD-similarity matrix across the seven cities.}
    \label{fig:fig4}
\end{figure}

Figure 5 displays the posterior $\lambda$ by city and software version under the GP-prior model. High-volume cells (San Francisco and Phoenix Version 10) show tight credible intervals consistent with their realized experience; sparse version cells widen by an order of magnitude. 

\begin{figure}
    \centering
    \includegraphics[width=0.7\linewidth]{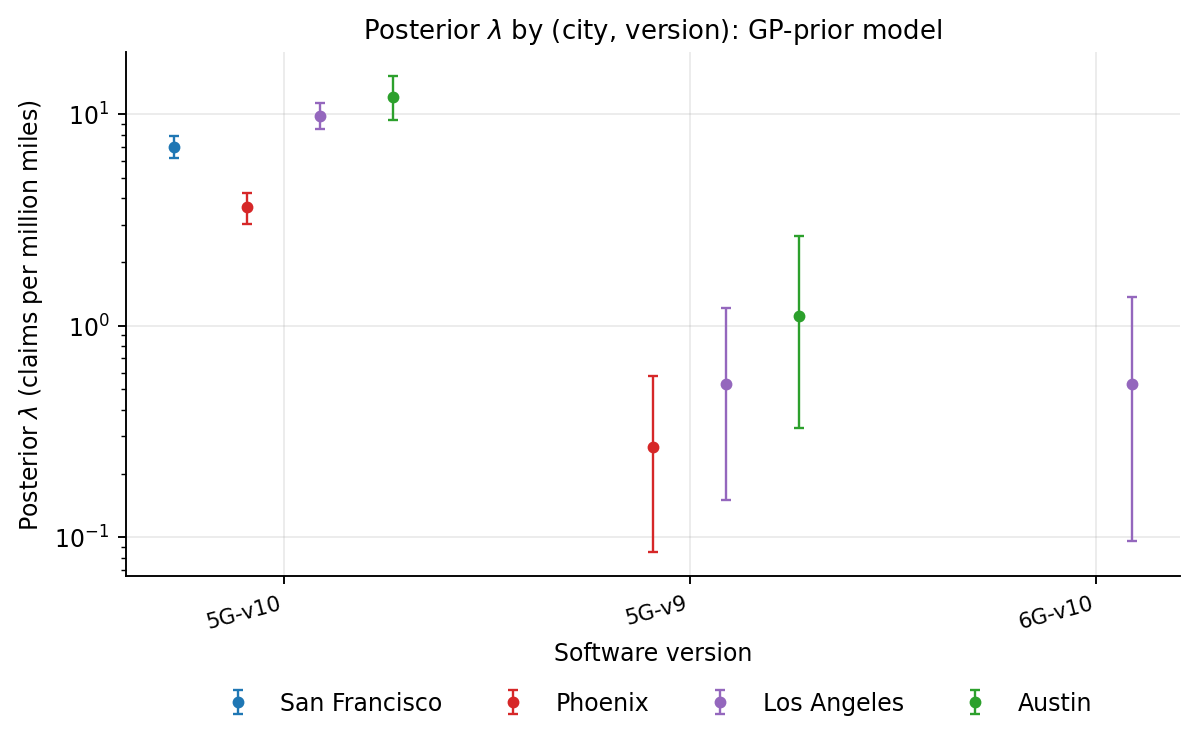}
    \caption{Posterior $\lambda$ by (city, software version) under the GP-prior model.}
    \label{fig:fig5}
\end{figure}

\subsection{Prospective Estimation for Hypothetical New Deployments}

To demonstrate the prospective value of the framework we construct hypothetical new-city deployments in Miami, Boston, and Denver. For each, we compute the ODD-embedding similarity to existing deployments, derive the implied prior on the city random effect, and report the prospective posterior frequency before any local experience accrues, via the conditional Gaussian formula of Section 5.3 (Appendix C). Table 4 reports the posterior median and 95 percent credible interval for each new city; Figure 6 visualizes them.

\begin{table}[htbp]
\centering
\caption{Prospective posterior predictive distribution of lambda for hypothetical new cities}
\begin{tabular}{llcccc}
\toprule
\textbf{City} & \textbf{Primary neighbor} & \textbf{Max similarity} & \textbf{Median} & \textbf{2.5\%} & \textbf{97.5\%} \\ 
\midrule
Miami  & San Francisco & 0.797 & 0.680 & 0.045 & 11.345 \\ 
Boston & San Francisco & 0.868 & 1.176 & 0.143 & 8.276 \\ 
Denver & Austin        & 0.760 & 0.981 & 0.203 & 3.935 \\ 
\bottomrule
\end{tabular}
\end{table}

\begin{figure}
    \centering
    \includegraphics[width=0.7\linewidth]{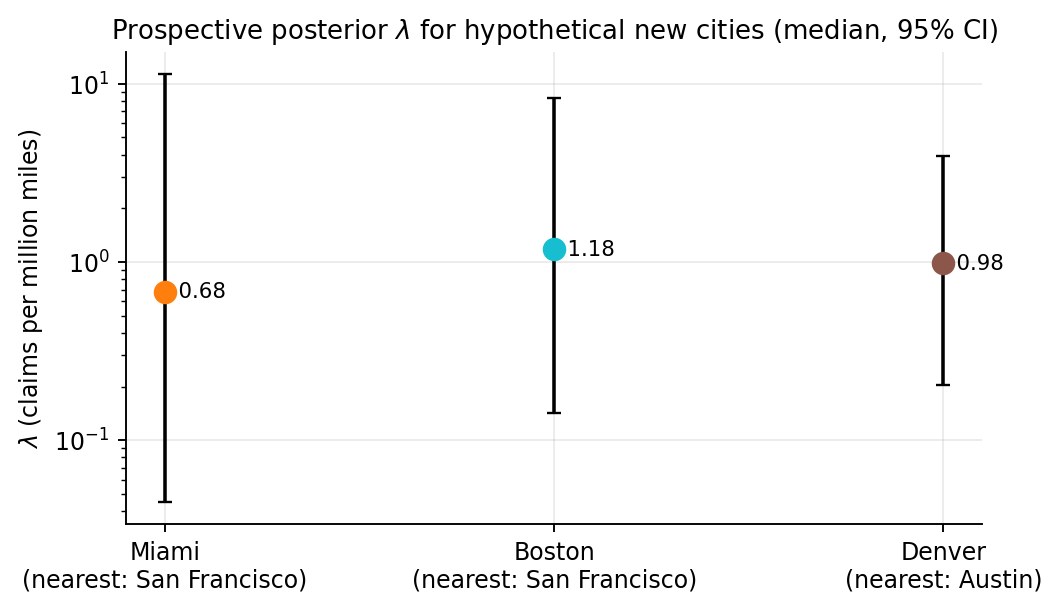}
    \caption{Posterior medians and 95 percent credible intervals for three hypothetical new deployments.}
    \label{fig:fig6}
\end{figure}

Denver (primary deployed neighbor Austin at S = 0.76, secondary Phoenix at S = 0.74) draws its prior from the sunbelt-arterial cluster: posterior median 0.981 crashes per million miles, 95 percent interval (0.203, 3.935). Boston (primary neighbor San Francisco at S = 0.87) draws from the dense-urban cluster: median 1.176, interval (0.143, 8.276). Miami (primary neighbor San Francisco at S = 0.80) has median 0.680 but the widest interval, (0.045, 11.345). The ordering of interval widths---Denver tightest, then Boston, then Miami---reflects both maximum embedding similarity and how informative the primary neighbor’s deployed experience is. The framework is designed to produce exactly this behavior: the more like a deployed city the new territory is, and the richer that deployed city’s experience, the tighter the prospective estimate.

The framework also supports principled updating as local experience accumulates. We illustrate with a hypothetical first-million-miles scenario for each prospective city under three experience levels---0, 1, and 3 observed crashes---chosen to span below-, at-, and above-prior outcomes on the SGO scale (where the prior medians are roughly 0.6 to 1.2 crashes per million miles). Showing the grid rather than a single arbitrary count makes the updating mechanism visible and keeps the demonstration from hinging on one Poisson draw. We update by importance-reweighting the prospective posterior draws: each draw $\lambda^{(s)}$ is reweighted by the Poisson likelihood of the hypothetical new observations, and the weighted ensemble is summarized as posterior median and 95 percent interval. Table 5 and Figure 7 report the result.

\begin{table}[h!]
\centering
\caption{Posterior update after a hypothetical first million miles of experience}
\renewcommand{\arraystretch}{1.2} 
\begin{tabular}{lccc}
\toprule
\textbf{City} & \textbf{Prior median} & \textbf{Post. median (0 / 1 / 3 crashes)} & \textbf{ESS (0/1/3)} \\ 
\midrule
Miami  & 0.680 & 0.304 / 0.753 / 2.130 & 2091 / 2320 / 1165 \\ 
Boston & 1.176 & 0.554 / 0.992 / 2.187 & 1867 / 2476 / 1701 \\ 
Denver & 0.981 & 0.646 / 0.957 / 1.758 & 2255 / 2793 / 1691 \\ 
\bottomrule
\end{tabular}
\end{table}

\begin{figure}
    \centering
    \includegraphics[width=0.75\linewidth]{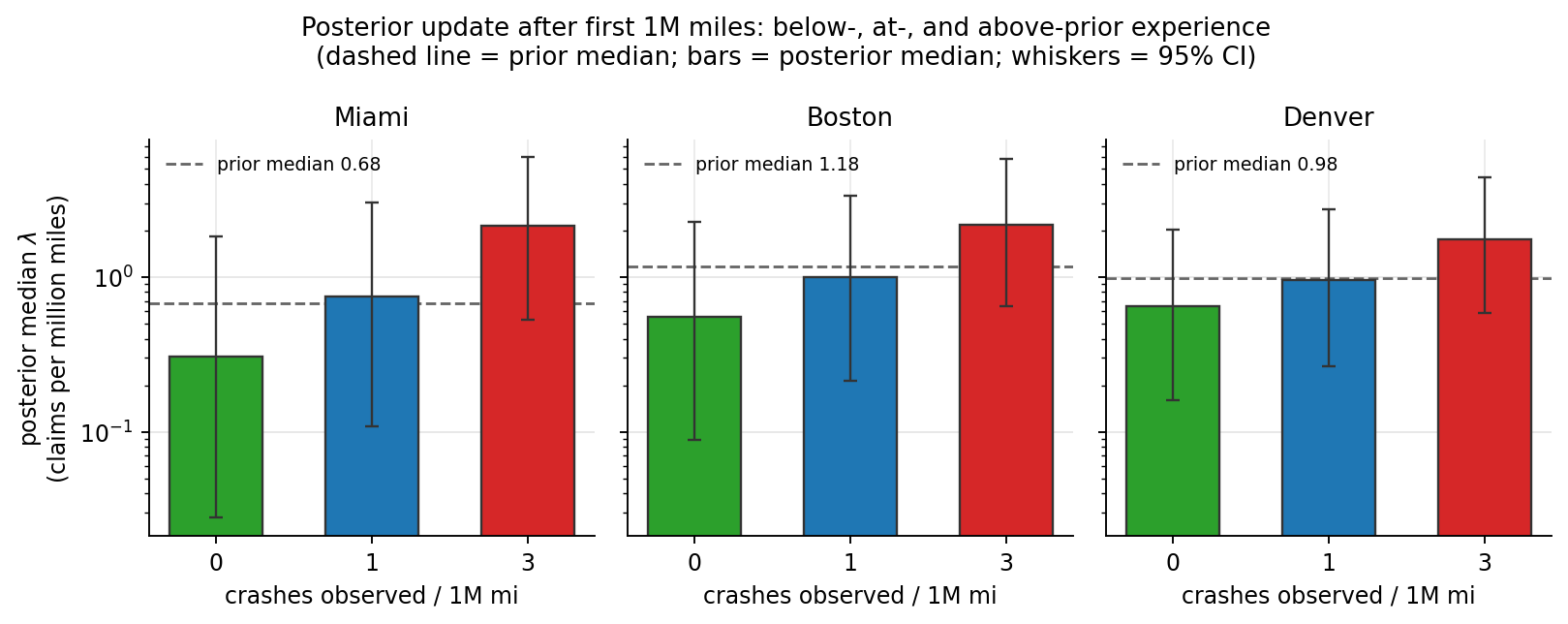}
    \caption{Posterior median expected frequency after a hypothetical first million miles}
    \label{fig:fig7}
\end{figure}

The Bayesian update behaves exactly as theory predicts. At one observed crash per million miles---roughly the at-prior outcome on the SGO scale---each posterior median sits close to its prior: Boston moves from 1.176 to 0.992, Miami from 0.680 to 0.753, Denver from 0.981 to 0.957. A favorable first million (zero crashes) pulls every posterior below its prior (Boston to 0.554, Miami to 0.304, Denver to 0.646), and an adverse first million (three crashes) pulls every posterior above it (Boston to 2.187, Miami to 2.130, Denver to 1.758). The magnitude of the move tracks prior width exactly as credibility theory requires: Miami, with the most diffuse prior, swings most across the three scenarios (0.304 to 2.130), while Denver, with the tightest prior, swings least (0.646 to 1.758). In every case the posterior is a coherent blend of one million miles of own experience and the cross-city prior.

\subsection{Comparison Against Baselines}

We evaluate whether adding cross-city correlation structure to the city random-effect prior improves out-of-sample predictive accuracy over an exchangeable city prior. All four random-effects models share the same fixed effects (intercept, six ODD covariates, software-version random effects, city-version interactions) and the same $\mathrm{HalfNormal}(0.5)$ hyperprior on the city scale parameter. The kernel is the sole difference: independent-RE assumes zero covariance between cities ($\alpha_i \perp \alpha_j$), while the GP models impose covariance $\sigma_c^2 S_{ij}$. Because $S$ is normalized to unit diagonal, the marginal prior variance of each city's random effect is identical under all models ($\sigma_c^2 = \tau_c^2$ at matched scale), so the comparison isolates only whether the off-diagonal entries of $S$ carry useful cross-city information. For the GP models the held-out city's posterior is the conditional Gaussian mean $\mathbf{s}^\top S^{-1} \boldsymbol{\alpha}_{\mathrm{train}}$---a kernel-weighted average of the training city posteriors; for independent-RE the conditional collapses to the prior mean (zero covariance implies zero conditional mean), so prediction reduces to pure shrinkage toward the grand mean. The question is therefore: do the off-diagonal entries of $S$ encode cross-city structure beyond what treating cities as exchangeable provides?

We compare five models through leave-one-city-out (LOCO) predictive log-likelihood: a single pool (no random effects), the hierarchical independent-RE model, and three GP-prior variants whose kernels are built from the Euclidean covariate distance, a one-dimensional log-HDV-frequency similarity, and the learned embedding. For each held-out city the remaining three cities fit each model; the held-out city’s aggregate crashes and exposure are scored under the resulting posterior predictive.

A single global pool predicts a held-out city badly, incurring a large Poisson penalty whenever the held city’s rate departs from the pooled mean. The random-effects models, which can shrink toward city-specific structure and borrow through covariates and the kernel, recover most of that loss. Table 6 reports the per-held-city and total LOCO log-likelihoods. Because held-out Poisson log-likelihoods are noisy at $K=4$ cities, the ordering across the four partial-pooling models varies across MCMC seeds; no kernel consistently dominates. Section 6.5 shows this is a granularity problem: four held-out cities carry too much per-realization noise for a single leave-one-out run to resolve that advantage at current data volumes.

\begin{table}[ht]
\centering
\caption{Leave-one-city-out predictive log-likelihood by model.}
\label{tab:loco}
\begin{tabular}{lrrrrr}
\toprule
\textbf{Held-out city} & \textbf{Single pool} & \textbf{Indep RE} & \textbf{GP Euclidean} & \textbf{GP HDV-only} & \textbf{GP learned} \\
\midrule
San Francisco & $-16.99$ & $-6.63$ & $-6.76$ & $-7.04$ & $-6.95$ \\
Phoenix       & $-86.94$ & $-5.78$ & $-5.88$ & $-5.87$ & $-5.95$ \\
Los Angeles   & $-13.36$ & $-7.62$ & $-7.99$ & $-8.27$ & $-7.43$ \\
Austin        & $-15.87$ & $-8.16$ & $-9.60$ & $-8.39$ & $-9.84$ \\
\midrule
\textbf{Total} & $-133.16$ & $-28.19$ & $-30.23$ & $-29.57$ & $-30.17$ \\
\bottomrule
\end{tabular}
\end{table}

\subsection{How Much Data Would Separate the Kernels? A Power Analysis}

The LOCO comparison establishes that no kernel separates from the others at current granularity, but it cannot say whether that is because the learned kernel carries no advantage or because the data are too thin to reveal one. These are different claims with different remedies, and they can be distinguished by simulation. We therefore ask: if city random effects genuinely follow the learned ODD-similarity covariance, how much claim volume would be required before the learned kernel’s out-of-sample advantage over the misspecified alternatives becomes statistically detectable?

We replicate the Section 6.4 evaluation on simulated data whose city log-effects are drawn from a multivariate normal with covariance proportional to the learned similarity matrix, so that the learned kernel is by construction correctly specified. Per-city exposure is swept so the expected total claim count ranges from a handful to several thousand; at each volume we draw 2000 replicate datasets and record the learned kernel's LOCO advantage over the alternatives. As specificity checks we repeat the exercise under independent and Euclidean-covariance null processes, under which a sound method must not favor the learned kernel. 

\begin{figure}[htbp]
    \centering
    \includegraphics[width=\linewidth]{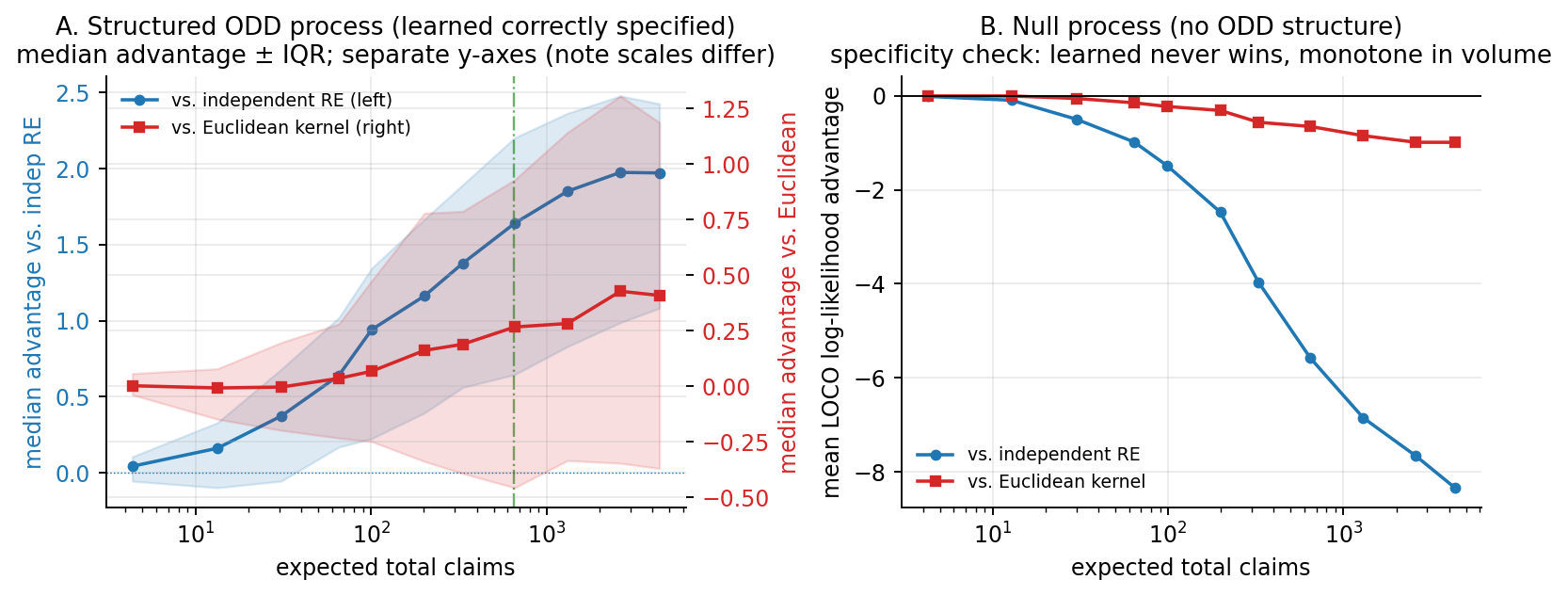}
    \caption{Forward-simulation power analysis. Panel A: under a data-generating process whose city effects follow the learned ODD-similarity structure, the mean leave-one-city-out log-likelihood advantage of the learned kernel over the independent-RE and Euclidean-kernel baselines, with two-standard-error bands, as a function of expected total claims (log scale). Panel B: under a null process with no ODD structure, the learned kernel never wins.}
    \label{fig:fig8}
\end{figure}

Under the structured process the learned kernel's median advantage rises with exposure. At the real-data volume of 648 claims (green dashed line) the median advantage reaches approximately +1.65 nats against independent-RE and +0.75 nats against the Euclidean kernel. Under either null process the learned kernel never attains a positive advantage, confirming specificity.

A single leave-one-city-out evaluation carries the full per-replicate spread, and with only four held-out cities that spread is large. At 648 claims the interquartile range of the learned-minus-independent-RE advantage spans roughly 0 to +2.5 nats (Panel A, shaded band), meaning roughly one quarter of single four-city realizations show no advantage for the learned kernel even at real-data volume. The limiting noise is the draw of four city effects, not the Poisson count, so a single LOCO cannot reliably resolve a modest kernel advantage regardless of claim volume. Section 6.4 is exactly one such four-fold draw; the tie it reports among the random-effects kernels is consistent with the power analysis.

The natural question is whether adding more deployed cities would resolve the comparison, and if so how many. We answer it directly with a second, complementary power analysis that holds per-city volume fixed (at a modest 40 claims per city) and instead sweeps the number of deployed cities K, recording how often a single LOCO run favors the learned kernel. The cities beyond the real deployed four are synthesized by sampling 32-dimensional ODD embeddings that match the learned seven-city similarity geometry (same leading subspace and per-axis spread, re-normalized to the unit sphere), so the learned kernel remains correctly specified by construction. Figure 9 plots the result. Detection power climbs steadily with K: reaching the conventional 80-percent level against the Euclidean kernel requires about twelve deployed cities. Under a null process with independent city effects the detection probability stays near or below chance at every K, confirming the test is specific. 

\begin{figure}
    \centering
    \includegraphics[width=0.7\linewidth]{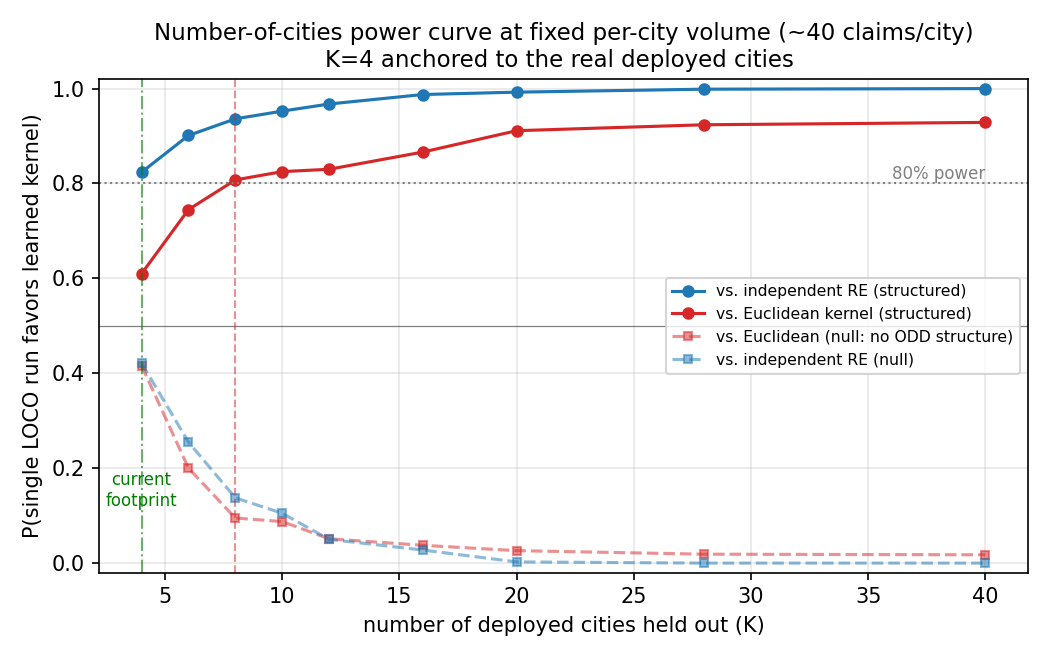}
    \caption{Number-of-cities power curve.}
    \label{fig:fig9}
\end{figure}

\section{Discussion}

\subsection{Implications for Ratemaking and Reserving}

The framework has direct implications for ADS ratemaking. The hierarchical credibility model provides a principled basis for filing rates in new geographies before local experience has accumulated, replacing ad hoc judgmental loadings with posterior credibility weights tied to observable ODD features. The ODD-similarity kernel makes the borrowing logic auditable: a regulator can ask which existing deployments inform the rate for a proposed new deployment, and the conditional Gaussian regression coefficients of Section 5.3 supply a numerical answer. Section 6.3 shows what such a filing might look like---for a hypothetical Boston deployment the prospective posterior median is 1.18 SGO crashes per million miles, drawn primarily from San Francisco experience through the 0.87 embedding similarity, with an explicit 95 percent credible interval; as the first million miles of Boston experience accumulate, the posterior updates in the appropriate direction at a Bayesian-coherent rate, neither overweighting the small sample nor ignoring it.

A natural extension is to couple frequency with severity into a full pure-premium model, and to broaden the random-effect granularity beyond city level\textemdash the H3 cell is the natural finest unit and the SimCLR embedding already exists at that resolution, making the challenge computational rather than conceptual.

\subsection{Limitations}

First, SGO is not a liability-claim proxy. SGO captures every crash involving an engaged ADS regardless of fault, whereas insurers adjudicate fault and record only the subset that generates a claim. The gap between the SGO crash rate (about 5.4 per million miles in our window) and the Di Lillo et al. (2024b) liability rate (about 0.43 per million miles) is roughly an order of magnitude. A practitioner using this framework for rate-filing purposes must supply a liability-claims numerator rather than raw SGO counts; the model’s structure accommodates either, but the interpretation of every $\lambda$ in Section 6 is an SGO-reportable-crash rate, not a liability rate.

Second, software-version granularity is limited in public data. After canonicalizing typographical variants, only three distinct versions are observable across 648 records, and the dominant label accounts for roughly 99 percent. At this granularity the city-by-version interaction $\delta_{c,v}$ is weakly identified even though the version main effect $\gamma_v$ is not. Practitioners with access to per-update mileage and claim data would observe substantially richer version structure.

Third, the framework addresses frequency, not severity. A complete pure-premium model would couple the hierarchical frequency model with a severity model---potentially using the same embedding apparatus to capture spatial variation in severity. We leave this extension to future work; the conditional-independence structure of standard frequency–severity models makes the combination straightforward in principle.

Fourth, the embedding may miss ADS-specific risk factors. It is trained on observable ODD features, so to the extent that ADS-specific risk differs from HDV-relevant risk---novel failure modes around construction zones, emergency vehicles, or particular intersection geometries — the embedding may not capture it. The contrastive framework readily accommodates additional features as the ADS-specific risk literature matures; the binding constraint is feature availability rather than methodology.

\section{Conclusion}

Automated Driving System deployments present a structural challenge to actuarial ratemaking: sparse exposure at the granularity that matters, shifting operational design domains, and a non-stationary insured risk. The retrospective benchmarking literature has established that ADS frequencies are substantially lower than HDV frequencies but has not addressed the prospective pricing problem insurers face as fleets expand into new geographies. This paper proposes a hierarchical Bayesian credibility framework that integrates classical credibility theory with a learned ODD-similarity metric, providing a principled basis for transferring crash experience across cities, software versions, and territories, demonstrated on NHTSA SGO data.

Two findings emerge from the empirical demonstration. First, with 648 crashes the city-aggregate Bühlmann–Straub credibility weights are moderate (0.12–0.46), confirming that the volume problem is not solved at the city aggregate and is sharper still at the per-(city, version, period) cell level. Second, partial pooling decisively outperforms no pooling; the learned kernel does not separate from simpler kernels in a single four-city evaluation, but a matched power analysis shows this reflects a granularity limit---the per-realization noise of four held-out cities swamps the modest mean advantage---not a verdict on the kernel's value. 

Future work will couple the frequency model with severity for full pure-premium estimation, develop an ADS-aware loss-development model for the faster claim emergence of sensor-equipped vehicles, and broaden the embedding feature set as ADS-specific risk factors mature.

\section*{References}
\begin{flushleft}
\hangindent=0.5in
\hangafter=1

Betancourt, M., \& Girolami, M. (2015). Hamiltonian Monte Carlo for hierarchical models. In S. K. Upadhyay, U. Singh, D. K. Dey, \& A. Loganathan (Eds.), Current Trends in Bayesian Methodology with Applications. CRC Press.

Bühlmann, H. (1967). Experience rating and credibility. ASTIN Bulletin, 4(3), 199–207.

Bühlmann, H., \& Straub, E. (1970). Glaubwürdigkeit für Schadensätze. Bulletin of the Swiss Association of Actuaries, 70, 111–133.

Bühlmann, H., \& Gisler, A. (2005). A Course in Credibility Theory and its Applications. Springer.

Chen, J. J., \& Shladover, S. E. (2024). Initial indications of safety of driverless automated driving systems. arXiv preprint arXiv:2403.14648.

Chen, T., Kornblith, S., Norouzi, M., \& Hinton, G. (2020). A simple framework for contrastive learning of visual representations. Proceedings of the 37th International Conference on Machine Learning, 1597–1607.

Chen, Y., Scanlon, J. M., Kusano, K. D., McMurry, T., \& Victor, T. (2024). Dynamic benchmarks: Spatial and temporal alignment for ADS performance evaluation. arXiv preprint arXiv:2410.08903.

Delong, Ł., Lindholm, M., \& Wüthrich, M. V. (2021). Gamma mixture density networks and their application to modelling insurance claim amounts. Insurance: Mathematics and Economics, 101, 240–261.

Di Lillo, L., Gode, T., Zhou, X., Atzei, M., Chen, R., \& Victor, T. (2024a). Comparative safety performance of autonomous- and human drivers: A real-world case study of the Waymo Driver. Heliyon, 10(14).

Di Lillo, L., Gode, T., Zhou, X., Scanlon, J., Chen, R., \& Victor, T. (2024b). Do autonomous vehicles outperform latest-generation human-driven vehicles? A comparison to Waymo's auto liability insurance claims at 25 million miles. Swiss Re / Waymo Technical Report.

Favarò, F. M., Nader, N., Eurich, S. O., Tripp, M., \& Varadaraju, N. (2017). Examining accident reports involving autonomous vehicles in California. PLoS One, 12(9), e0184952.

Gelman, A., Jakulin, A., Pittau, M. G., \& Su, Y.-S. (2008). A weakly informative default prior distribution for logistic and other regression models. Annals of Applied Statistics, 2(4), 1360–1383.

Khosla, P., Teterwak, P., Wang, C., Sarna, A., Tian, Y., Isola, P., Maschinot, A., Liu, C., \& Krishnan, D. (2020). Supervised contrastive learning. Advances in Neural Information Processing Systems, 33, 18661–18673.

Klugman, S. A., Panjer, H. H., \& Willmot, G. E. (2012). Loss Models: From Data to Decisions (4th ed.). Wiley.

Kusano, K. D., Scanlon, J. M., Chen, Y. H., McMurry, T. L., Chen, R., Gode, T., \& Victor, T. (2024). Comparison of Waymo rider-only crash data to human benchmarks at 7.1 million miles. Traffic Injury Prevention, 1–12.

National Highway Traffic Safety Administration. (2026). Standing General Order on Crash Reporting. NHTSA Public Data Releases.

Phan, D., Pradhan, N., \& Jankowiak, M. (2019). Composable effects for flexible and accelerated probabilistic programming in NumPyro. arXiv preprint arXiv:1912.11554.

Richman, R. (2021). AI in actuarial science: A review of recent advances—Part 1 and Part 2. Annals of Actuarial Science, 15(2), 207–290.

Scanlon, J. M., Kusano, K. D., Fraade-Blanar, L. A., McMurry, T. L., Chen, Y. H., \& Victor, T. (2024a). Benchmarks for retrospective automated driving system crash rate analysis using police-reported crash data. Traffic Injury Prevention, 25(sup1), S51–S65.

Scanlon, J. M., Teoh, E. R., Kidd, D. G., Kusano, K. D., Bärgman, J., Chi-Johnston, G., Di Lillo, L., Favaro, F., Flannagan, C., Liers, H., Lin, B., Lindman, M., McLaughlin, S., Perez, M., \& Victor, T. (2024b). RAVE checklist: Recommendations for overcoming challenges in retrospective safety studies of automated driving systems. Traffic Injury Prevention

Uber Engineering. (2018). H3: A hexagonal hierarchical geospatial indexing system. https://h3geo.org/

Wüthrich, M. V., \& Buser, C. (2025). Data Analytics for Non-Life Insurance Pricing. Lecture notes, ETH Zurich.

Wüthrich, M. V., \& Merz, M. (2019). Editorial: Yes, we CANN! ASTIN Bulletin, 49(1), 1–3.

\end{flushleft}

\appendix

\definecolor{codebg}{HTML}{F8F8F8}
\definecolor{kw}{HTML}{0060A0}
\definecolor{cmt}{HTML}{408040}
\definecolor{str}{HTML}{C04000}
\lstdefinestyle{python}{
  language=Python,
  basicstyle=\ttfamily\small,
  backgroundcolor=\color{codebg},
  keywordstyle=\color{kw}\bfseries,
  commentstyle=\color{cmt}\itshape,
  stringstyle=\color{str},
  showstringspaces=false,
  breaklines=true,
  frame=single,
  framerule=0.4pt,
  rulecolor=\color{gray!40},
  aboveskip=6pt,
  belowskip=6pt,
  xleftmargin=8pt,
  xrightmargin=4pt,
  numbers=none,
  morekeywords={numpyro,dist,MCMC,NUTS,jnp,deterministic,sample},
}

\newcommand{\R}{\mathbb{R}}
\newcommand{\N}{\mathcal{N}}
\newcommand{\E}{\mathbb{E}}

\newcommand{\Sdep}{S_{\mathrm{dep}}}
\newcommand{\adep}{\alpha_{\mathrm{dep}}}
\newcommand{\cs}{c^{*}}

\renewcommand{\thesection}{\Alph{section}}
\renewcommand{\thesubsection}{\Alph{section}.\arabic{subsection}}
\renewcommand{\thetable}{\Alph{section}.\arabic{table}}
\counterwithin{table}{section}

\section{NumPyro Implementation}
\label{app:numpyro}

The following is the NumPyro implementation of the hierarchical model of Section~4. The GP-prior extension of Section~5.3 follows. The complete pipeline---SGO fetch and version canonicalization, embedding training, MCMC inference, and figure generation---is reproducible by running:

\begin{verbatim}
python -m src.fetchers.sgo_fetcher --out data --operator 'Waymo LLC'

python src/embedding.py --data-dir data --out results/embedding

python src/run_analysis.py --data-dir data \
       --embed-dir results/embedding --out results

python src/make_figures.py --data-dir data \
       --embed-dir results/embedding \
       --results-dir results --out figures
\end{verbatim}

\subsection*{Baseline hierarchical model}

\begin{verbatim}
import numpyro, jax.numpy as jnp
import numpyro.distributions as dist
from numpyro.infer import MCMC, NUTS

def ads_credibility_model(city_idx, version_idx, X, exposure,
                          claims=None, n_cities=4, n_versions=3,
                          n_features=6):

    beta0 = numpyro.sample('beta0', dist.Normal(0., 2.5))
    beta  = numpyro.sample('beta',
        dist.Normal(jnp.zeros(n_features), 0.5))   # regularizing prior (Sec 4.1)

    tau_c  = numpyro.sample('tau_c',  dist.HalfNormal(0.5))
    tau_v  = numpyro.sample('tau_v',  dist.HalfNormal(0.5))
    tau_cv = numpyro.sample('tau_cv', dist.HalfNormal(0.3))

    # non-centered random effects
    a_raw = numpyro.sample('alpha_raw', dist.Normal(jnp.zeros(n_cities), 1.))
    g_raw = numpyro.sample('gamma_raw', dist.Normal(jnp.zeros(n_versions), 1.))
    d_raw = numpyro.sample('delta_raw',
        dist.Normal(jnp.zeros((n_cities, n_versions)), 1.))

    alpha = numpyro.deterministic('alpha', tau_c * a_raw)
    gamma = numpyro.deterministic('gamma', tau_v * g_raw)
    delta = numpyro.deterministic('delta', tau_cv * d_raw)

    log_lambda = (beta0 + X @ beta + alpha[city_idx]
                  + gamma[version_idx] + delta[city_idx, version_idx])

    numpyro.sample('claims',
                   dist.Poisson(jnp.exp(log_lambda) * exposure),
                   obs=claims)
\end{verbatim}

\subsection*{GP-prior extension}

The GP-prior extension replaces the independent \texttt{alpha\_raw} block
with a multivariate-normal draw whose covariance is the trained similarity
matrix $\mathbf{S}$, via a Cholesky factor $\mathbf{L}$ with
$\mathbf{L}\mathbf{L}^{\top} = \mathbf{S}$:

\begin{verbatim}
def ads_credibility_gp_model(city_idx, version_idx, X, exposure, L_chol,
                             claims=None, n_cities=4, n_versions=3,
                             n_features=6):

    beta0   = numpyro.sample('beta0', dist.Normal(0., 2.5))
    beta    = numpyro.sample('beta', dist.Normal(jnp.zeros(n_features), 0.5))
    sigma_c = numpyro.sample('sigma_c', dist.HalfNormal(0.5))
    tau_v   = numpyro.sample('tau_v',   dist.HalfNormal(0.5))
    tau_cv  = numpyro.sample('tau_cv',  dist.HalfNormal(0.3))

    z_alpha = numpyro.sample('z_alpha', dist.Normal(jnp.zeros(n_cities), 1.))
    alpha   = numpyro.deterministic('alpha', sigma_c * (L_chol @ z_alpha))

    g_raw = numpyro.sample('gamma_raw', dist.Normal(jnp.zeros(n_versions), 1.))
    d_raw = numpyro.sample('delta_raw',
        dist.Normal(jnp.zeros((n_cities, n_versions)), 1.))

    gamma = numpyro.deterministic('gamma', tau_v * g_raw)
    delta = numpyro.deterministic('delta', tau_cv * d_raw)

    log_lambda = (beta0 + X @ beta + alpha[city_idx]
                  + gamma[version_idx] + delta[city_idx, version_idx])

    numpyro.sample('claims',
                   dist.Poisson(jnp.exp(log_lambda) * exposure),
                   obs=claims)
\end{verbatim}

Software versions used in the reported runs: Python~3.12, NumPyro~0.18,
JAX~0.6, PyTorch~2.5, NumPy~2.1, ArviZ~0.20, scikit-learn~1.6,
Pandas~2.3.

\section{Derivation of the B\"{u}hlmann--Straub Limit Case}
\label{app:bs}

We derive the B\"{u}hlmann--Straub credibility formula from the hierarchical
Bayesian Poisson model of Section~4 in the simplified case of no covariates,
no software-version effect, and no city--version interaction:
\[
  N_c \mid \lambda_c \;\sim\; \mathrm{Poisson}(\lambda_c \cdot E_c),
  \qquad
  \log \lambda_c = \beta_0 + \alpha_c,
  \qquad
  \alpha_c \;\sim\; \mathcal{N}(0,\,\tau^2).
\]
Take $\beta_0$ and $\tau$ as fixed. The log-likelihood contribution of
city $c$ is
\[
  \ell_c(\alpha_c)
    = N_c\,(\beta_0 + \alpha_c)
      - E_c\,\exp(\beta_0 + \alpha_c)
      + \text{const}.
\]
Applying a Laplace approximation around the posterior mode $\hat\alpha_c$,
the first-order condition is
\[
  N_c - E_c\,\exp(\beta_0 + \hat\alpha_c) - \hat\alpha_c/\tau^2 = 0.
\]
Replacing $\exp(\beta_0+\hat\alpha_c)$ by the plug-in
$\hat\lambda_c^{r} = N_c/E_c$ (exact at the MLE under the
reparameterisation $\lambda_c = \exp(\beta_0+\alpha_c)$) gives
\[
  \hat\alpha_c
    = \frac{E_c \cdot \hat\lambda_c^{r} \cdot \tau^2}
           {E_c \cdot \hat\lambda_c^{r} \cdot \tau^2 + 1}
      \cdot \bigl(\log\hat\lambda_c^{r} - \beta_0\bigr)
    = Z_c \cdot \bigl(\log\hat\lambda_c^{r} - \beta_0\bigr),
\]
with
\[
  Z_c = \frac{w_c}{w_c + K},
  \qquad
  w_c = E_c \cdot \hat\lambda_c^{r} = N_c,
  \qquad
  K = 1/\tau^2.
\]
As $\tau\to\infty$, $Z_c\to 1$ and the posterior collapses to own
experience; as $\tau\to 0$, $Z_c\to 0$ and it collapses to the grand mean.
The derivation depends on the Laplace approximation, which is exact for
Gaussian posteriors and accurate for Poisson posteriors with moderate
expected counts; the full Bayesian computation in Section~6 does not rely
on it. The derivation generalises to the model with covariates by carrying
$\mathbf{x}^{\top}\!\boldsymbol\beta$ through the calculation, with
$\hat\lambda_c^{r}$ replaced by the residual rate after subtracting the
fixed-effect contribution. $\square$

\section{Posterior Predictive for a Held-Out City}
\label{app:gp_pred}

We derive the conditional Gaussian for the new-city random effect under
the GP-prior model. Let $c$ index the $C$ deployed cities and $c^*$ a
single new city. Stack the random effects
$(\boldsymbol\alpha_{\mathrm{dep}},\alpha_{c^*})^{\top}
\sim \mathcal{N}(\mathbf{0},\,\sigma^2\boldsymbol\Sigma)$,
where $\boldsymbol\Sigma$ is the $(C+1)\times(C+1)$ joint similarity
matrix with blocks $\mathbf{S}_{\mathrm{dep}}$ (deployed),
$\mathbf{s}$ (new-city-to-deployed), and $s_*$ (new-city
self-similarity). The standard partition formula gives
\begin{align*}
  \alpha_{c^*} \mid \boldsymbol\alpha_{\mathrm{dep}},\,\sigma
    &\;\sim\; \mathcal{N}(\mu_{\mathrm{cond}},\,\sigma^2_{\mathrm{cond}}),
  \\
  \mu_{\mathrm{cond}}
    &= \mathbf{s}^{\top}\mathbf{S}_{\mathrm{dep}}^{-1}
       \boldsymbol\alpha_{\mathrm{dep}},
  \\
  \sigma^2_{\mathrm{cond}}
    &= \sigma^2\!\left(s_* - \mathbf{s}^{\top}
       \mathbf{S}_{\mathrm{dep}}^{-1}\mathbf{s}\right).
\end{align*}
The conditional mean is a kernel regression of the deployed-city random
effects onto the new city, with weights
$\mathbf{s}^{\top}\mathbf{S}_{\mathrm{dep}}^{-1}$---the standard
Gaussian-process predictive weights. The conditional variance is zero when
$c^*$ is perfectly similar to a deployed city and equals the marginal prior
variance $\sigma^2$ when $c^*$ is orthogonal to the deployed set. In
practice we add a small jitter ($10^{-6}\cdot\mathbf{I}$) to
$\mathbf{S}_{\mathrm{dep}}$ before factorisation for numerical stability.
The posterior predictive of $\lambda_{c^*}$ follows by drawing
$(\boldsymbol\alpha_{\mathrm{dep}},\sigma)$ from the MCMC chain, drawing
$\alpha_{c^*}$ from the conditional Gaussian, and computing
$\lambda_{c^*} = \exp(\beta_0 + \mathbf{x}_{c^*}^{\top}\!\boldsymbol\beta
+ \alpha_{c^*})$ with $(\beta_0,\boldsymbol\beta)$ from the same draw.
$\square$

\section{Data Sources and Fetcher Pipeline}
\label{app:data}

\subsection{ADS crash counts (NHTSA SGO)}
\label{app:data:sgo}

Fetched from the NHTSA Standing General Order 2021-01 ADS incident-report file (NHTSA, 2026). We retain Waymo LLC reports with ``Verified Engaged'' status, deduplicate to the latest report version per incident identifier, map the incident municipality to one of four study metros, and canonicalise typographical variants of the version strings. The result is 648 crashes spanning June~2025--April~2026 across four deployed metros (San Francisco 254,
Phoenix 134, Los Angeles 186, Austin 74) and three canonical software versions, distributed across five quarterly periods.

\subsection{ADS exposure (operator disclosure)}
\label{app:data:exposure}

Derived from Waymo's published mileage milestones: 170.7~million
rider-only miles through December~2025 across the four metros (Waymo
Safety Impact hub) and the 200-million-mile milestone in February~2026.
Interpolating to the SGO window and allocating by each metro's share of
Waymo's published cumulative rider-only miles through December~2025 (San
Francisco 31.4\%, Phoenix 40.2\%, Los Angeles 22.2\%, Austin 6.3\%) yields
approximately 116~million four-city rider-only miles (San Francisco~36.37,
Phoenix~46.62, Los Angeles~25.72, Austin~7.29). These are estimates; a
sensitivity range of 100--130~million total yields SGO frequencies of
4.8--6.3 per million miles.

\subsection{ODD cell features (OSM, ACS, FARS)}
\label{app:data:odd}

5,239 H3 resolution-8 cells across seven cities (approximately 749 per
city). Road network from the OpenStreetMap Overpass API; demographics from
the Census ACS 5-year tabulations spatially joined to H3 cells; crash
density from NHTSA FARS 2018--2022; urbanised-area VMT from FHWA Highway
Statistics HM-72. The contrastive embedding is trained on this layer and is
independent of the SGO numerator.

\end{document}